\documentclass{ectj}
\usepackage[T1]{fontenc}
\usepackage{lmodern}

\usepackage{amsfonts,amssymb,graphics,epsfig,verbatim,bm,latexsym,amsmath,url,amsbsy, subcaption, booktabs}
\usepackage[ruled,vlined]{algorithm2e}
\usepackage[dvipsnames]{xcolor} 
\newtheorem{theorem}{Theorem}
\newtheorem{assumption}{Assumption}

\newtheorem{corollary}{Corollary}

\newtheorem{definition}{Definition}

\renewcommand{\thesection}{\arabic{section}}
\renewcommand{\theequation}{\arabic{section}.\arabic{equation}}
\renewcommand{\thetheorem}{\arabic{section}.\arabic{theorem}}

\renewcommand{\thecorollary}{\arabic{section}.\arabic{corollary}}

\newcommand{\parencite}[1]{(\citeauthor{#1}, \citeyear{#1})}

\received{...}
\accepted{...}
\volume{21}

\title[BPR for Environmental Prediction]{Bagged Polynomial Regression:\\
With Application to Environmental Prediction}

\author[Klosin and Vives-i-Bastida]{Sylvia~Klosin$^{\dagger}$ and
                Jaume~Vives-i-Bastida$^{\ddagger}$\thanks{We thank Dalia Ghanem, Vitor Hadad, Max Kasy, Whitney Newey, James Sayre, and Stefan Wager for helpful comments. All code for the project can be found at github.com/klosins/bpr.}}

\address{$^{\dagger}$Department of Agricultural and Resource Economics, UC Davis, Davis, CA 95616, USA.}

\email{sklosin@ucdavis.edu}

\address{$^{\ddagger}$Stanford Graduate School of Business, Stanford, CA 94305, USA.}
\email{vives@stanford.edu}

\def\AmSTeX{$\cal A$\kern-.1667em\lower.5ex\hbox{$\cal M$}\kern-.125em
    $\cal S$-\TeX}
\def\BibTeX{{\rm B\kern-.05em{\sc i\kern-.025em b}\kern-.08em
    T\kern-.1667em\lower.7ex\hbox{E}\kern-.125emX}}

\begin{document}
\setcounter{footnote}{1}

    \begin{abstract}

        Climate and environmental applications increasingly rely on high-dimensional prediction from remote sensing and other scientific data. Neural networks (NN) can deliver strong accuracy in these settings, but they are often hard to audit and hard to align with domain knowledge. As an alternative, we propose bagged polynomial regression with random projections (BPR), an econometrics-native ensemble that averages many regularized low-degree polynomial models fit on randomly selected covariate groups. We provide novel finite-sample and asymptotic risk bounds and show how covariate partitioning can improve rates for smooth target functions by controlling dictionary basis growth. Rate improvements may be particularly relevant for the estimation of marginal effects. In an application to satellite-based crop classification using optical and radar imagery, BPR matches NN accuracy while remaining straightforward to diagnose. We provide practical transparency tools, coefficient summaries and partial-dependence diagnostics, that show BPR captures intuitive feature relationships that NNs do not.


    \end{abstract}


\section{Introduction}
Deep neural networks (NN) have become ubiquitous in environmental economic prediction problems. Examples include flood detection \parencite{Patel2023Floods}, wildfire-smoke forecasts used to inform warnings \parencite{GellmanWallsWibbenmeyer2023SmokeWelfare}, and crop-type and yield prediction used for insurance and food-security early warning \parencite{KhakiPhamWang2021YieldNet,KeaneNeal2020EctJ}. In all of these settings, prediction is used to inform welfare calculations and policy design.\footnote{Emerging governance regimes (such as the EU AI Act) are raising the bar for transparency, documentation, and interpretability of these predictive algorithms.} As a result, predictive models must be both \emph{accurate} and \emph{legible} to domain scientists and policymakers.

NNs excel along the first of these dimensions - they are very accurate. This feat is possible by training very large (over-parametrized) models. However, their internal representations are difficult to interrogate; we show that common interpretable tools can be fragile in environmental settings. This paper proposes Bagged Polynomial Regression with random projections (BPR) and Partitioned Polynomial Regression (PPR), as econometrics-native alternatives that we show (i) come with formal theoretical guarantees, (ii) deliver competitive predictive performance in a high-dimensional environmental remote-sensing application, and (iii) retain a transparent, coefficient-based structure, that supports interpretability and leads to better marginal effect estimates than more complex models.

For the theoretical contribution, we build on the econometrics literature on series and polynomial regression methods, that are transparent, but adapt them for prediction in high dimensional settings. The central barrier is that the number of dictionary basis terms can explode with the covariate dimension, leading to finite-sample estimation errors shrinking too slowly to be practically useful. Our main results deliver finite-sample and asymptotic $L_2$ risk bounds for a class of partitioned series estimators (PPR), in which polynomial features are constructed group-wise after partitioning the covariates into blocks. The bounds make the estimation--approximation tradeoff explicit: partitioning reduces the effective basis complexity (and thus estimation error) while incurring a controlled loss in approximation error. We further characterize how the choice of partition size governs this tradeoff and provide conditions under which an intermediate level of partitioning is rate-improving, which in turn motivates our BPR estimator that averages polynomial base models fit on randomly selected feature groups. As a byproduct of our main theoretical results we also show that the estimation rates for marginal effects (Average Partial Derivatives) can be faster when the derivatives of the dictionary function are sparsely supported, further motivating models with less complex marginal structures when the researcher's objective is interpretability.

Second, we show that BPR can match state-of-the-art black-box predictors in an environmental economics application. Our main application is multi-class crop-type mapping from  satellite imagery (RapidEye optical and UAVSAR radar) near Winnipeg, Canada. The central prediction task is the full seven-class crop classification problem,\footnote{The crops in the data are corn, peas, canola, soybeans, oats, wheat, and broadleaf. } which is the most demanding setting in this application because it requires discriminating among multiple agronomically distinct but spectrally similar crop types. Accurate crop-type maps are important inputs to agricultural decision support and policy implementation (e.g., monitoring and compliance in subsidy programs) \parencite{SchmedtmannCampagnolo2015ReliableCropID}. When classifying all seven crops, BPR achieves high out-of-sample accuracy (test accuracy $0.997$), competitive with commonly reported neural-network baselines for this dataset (around $0.992$). We also demonstrate, in the Online Appendix, strong performance on a standard vision benchmark (MNIST): BPR attains close to state-of-the-art accuracy. In both settings, BPR achieves high accuracy while using substantially fewer parameters than alternative highly overparameterized models.

Third, we demonstrate that BPR can be more interpretable than NN alternatives. Unlike fully black-box architectures, BPR yields readable coefficients\footnote{BPR is comparatively transparent because it is built from regularized generalized linear models with an explicit coefficient-based representation, so the role of scientifically meaningful covariates can be audited using standard econometric summaries.} (levels, interactions, and polynomial contrasts), enabling alignment with scientific priors. This is helpful for environmental and climate-economics settings, where researchers often have domain knowledge about which variables should matter and the direction of their marginal effects\footnote{For example, in agricultural remote sensing, canopy biophysics implies that red--NIR contrast and soil-adjusted vegetation indices track vegetation density and condition, and applied studies consistently find these features predictive for crop identification and yield; see \citet{Taylor2010BandsPoppy,Jia2011SpectralPoppy,Waine2014ImprovingOpiumYield,Ghosh2003MaizeLAI}.}.  In particular, we consider two interpretability diagnostics commonly used by applied researchers (i) covariate coefficient summary tables and (ii) partial-dependency plots (PDPs).  We focus on the variable Normalized Difference Vegetation Index (NDVI), a measure of greenness of an image, as an example. For the covariate summary tables, we average NDVI’s coefficient across the bags that include NDVI, yielding a single signed, interpretable summary that is an estimate of the true Average Partial Derivative (marginal effect). This kind of coefficient-based audit is natural for BPR but not for neural networks, which may require numerical approximations. For the second diagnostic, PDP curves, we trace how predicted classification probabilities change with NDVI. Strikingly, in our application the NN PDP is essentially flat, even though NDVI is highly predictive,  while both the coefficient summary (APD) and PDP plot for BPR reveal a clear negative relationship, in line with the scientific priors. This matters because NDVI is a canonical, scientifically meaningful vegetation signal; BPR makes its role visible, letting researchers directly audit how changes in NDVI translate into changes in predicted crop probabilities.

\paragraph{Relation to the literature.}
This paper connects three literatures. First, we contribute to the series/sieve regression tradition that studies $L_2$ risk for approximating smooth functions with polynomial bases \parencite{newey1997convergence,stone1982optimal}. Building on \citet{belloni2015}, we use concentration results for random matrices by \parencite{rudelson2007sampling} to derive sharper $L_2$ bounds and new finite-sample rates for a broad class of series estimators. Second, we relate these results to ideas from ensemble learning and random subspaces by analyzing partitioned (group-wise) feature maps and characterizing how partition size governs the estimation--approximation tradeoff. This complements existing work on structured or partitioned nonparametric models, including additive and partitioning estimators \parencite{andrews1990additive,cattaneo2013optimal,khosravi2019non, iterForestsYu}. Third, we speak to the growing use of black-box prediction tools in environmental and agricultural economics by providing a competitive but more transparent alternative for settings such as floods, wildfire smoke, and crop classification \parencite{Patel2023Floods,GellmanWallsWibbenmeyer2023SmokeWelfare,KhakiPhamWang2021YieldNet,KeaneNeal2020EctJ}. Finally, our results complement recent comparisons of neural networks and polynomial methods \parencite{emschwiller2020neural,cheng2018polynomial} by providing theory and evidence for when an ensemble of low-degree polynomial models can match black-box performance while remaining interpretable.
\paragraph{Notation} In what follows we use $E$ to denote the expectation operator and $\mathbb{E}$ to denote the empirical expectation operator such that $\mathbb{E}_n[f(x)] = 1/n\sum_{i=1}^n f(x_i)$. Furthermore, unless stated otherwise we denote the $\ell_2$ norm for a vector by $\| \cdot \|$, the operator norm for a matrix $\| \cdot \|_{op}$ and let $a \lor b = \max\{a,b\}$. Finally, we write $ a \lesssim b$ to mean $a \leq Cb$ for some fixed constant $C>0$, $ a \lesssim_P b$ to mean $a = O_p(b)$ and $a\asymp b$ to mean $a$ is asymptotic to $b$. All proofs are provided in the Appendix and the Online Appendix.

\section{Theoretical results}
We derive asymptotic and finite-sample rates for sequences of models indexed by sample size $n$ that satisfy the following sampling assumption.

\begin{assumption}[Sampling model]
\label{eq:assump_sample_model}
For each $n$, random vectors $\{(y_i, x_i')\}_{i=1}^n$ are $i.i.d.$ and given by the series regression model
\begin{equation}
    y_i = g(x_i) +\epsilon_i,\quad E[\epsilon_i|x_i]=0,\quad x_i\in\mathcal{X}\subset \mathbb{R}^d,\quad i=1,\dots, n,
\end{equation}
where $y_i\in \mathbb{R}$ is the response variable, $x_i$ are the basic features in some bounded set $\mathcal{X}$, $\epsilon_i$ is a noise term and $g: \mathcal{X} \to \mathbb{R}$ is the conditional expectation function belonging to an arbitrary function class $\mathcal{G}$.
\end{assumption}

Given the sampling process, we focus on the set of models in which $g$ is approximated by linear forms $p(x)' b$, where $p(x):\mathbb{R}^d \to \mathbb{R}^k$ is a tensor operator that generates polynomial features, with $k$ being the total number of features generated from the basic features $x$. For an iid sample $\{y_i, x_i'\}$, $i = 1,\dots, n$, we estimate the series regressors by minimizing the statistical risk for the square loss function (i.e. the least squares problem):

\begin{equation}
\beta_g
:= \arg\min_{b \in \mathbb{R}^k}
\; E\!\left[\bigl(g(x_i) - p(x_i)' b\bigr)^2\right].
\end{equation}

\noindent where $\beta_g$ is the least squares estimate for a given conditional mean function $g$. This follows the setup in \cite{belloni2015} and the standard framework in the series regression literature (see \cite{newey1997convergence} or \cite{andrews1990additive}). A key object in this literature is the approximation error $r_g(x)$ for features $x$ and target function $g$, defined as
\begin{equation}
    r_g(x) = g(x) - p(x)'\beta_g.
\end{equation}
\noindent We can then rewrite the sampling model as linear regression model
\begin{equation}
    y_i = p_i'\beta + u_i, \quad E[u_ix_i] = 0,\quad u_i = r_i + \epsilon_i,
\end{equation}
and define the standard least squares projection estimator:
\begin{equation}
    \hat{\beta} = \arg\min_{b \in \mathbb{R}^k} \mathbb{E}_n \big[(y_i - p_i'b)^2 \big] = \mathbb{E}_n[p_ip_i']^{-1}\mathbb{E}_n[p_iy_i].
\end{equation}

Given the regression model we can decompose the error in estimating the target function $g(x)$ in two components, an estimation error and an approximation error
\begin{equation}
    \hat{g}(x) - g(x) = \underbrace{p(x)'(\hat{\beta} - \beta)}_{\text{estimation error}}\quad - \underbrace{r(x).}_{\text{approximation error}}
\end{equation}
Intuitively, the richer the polynomial embedding the smaller the approximation error will be, however this may come at the cost of a larger estimation error as the number of model parameters to estimate increases. Our theoretical results will show how to trade off these two errors in models in which we can constrain the feature map $p(x)$ to generate polynomial features for partitions of the feature space rather for all features. Observe that this trade off and our finite-sample results go beyond the standard bias-variance trade-off that only considers the estimation error. Even if the model optimally trade offs bias and variance to control the estimation error, it may still be beneficial to reduce or increase the richness of the feature map depending on the approximation error.

An important quantity to understand the role of the estimation and approximation errors when estimating the target function $g$ is
\begin{equation}
    \xi_k = \sup_{x\in \mathcal{X}} \|p(x)\|.
\end{equation}
When we have multiple basic features, and $\mathcal{X}$ is multi-dimensional, building a series of polynomial degree $J$ for each basic feature and then constructing all interactions leads to $k = \sum_{b=1}^{d} {d\choose b} J^b = (J+1)^d -1$ total features\footnote{This number can be thought as an upper bound on the amount of polynomial features built in practice, for instance with $d=2$ and $J=2$ such that $x = (x_1,x_2)$ it means we build \textit{all} interactions $\{x_1,x_1x_2,x_1x_2^2, x_1^2, x_1^2x_2, x_1^2x_2^2,x_2,x_2^2\}$. Our results, however, are easily amenable to other constructions of the polynomial embedding.}. Furthermore, when $\mathcal{X}$ is bounded (as required by assumption \ref{eq:assump_sample_model}), it can be shown that $\xi_k \precsim k$, which explodes exponentially in $d$, the dimension of the feature space $\mathcal{X}$. This is the main problem with polynomial regression that we motivated in the introduction. As $J$ and $d$ grow, $k$ becomes very large. For example, for $J=2$ and $d=40$, not a very high dimensional setting, $k$ is already of the order of $10^{19}$. On one hand, this leads to computational intractability as the OLS method requires inverting a matrix of size $k\times k$, and on the other hand, given that the $L^2$ convergence rate will depend on $k$ and require $k\to \infty$, as we will show in Theorem \ref{theorem:l_2_rates}, this means that the sample size necessary for convergence will also have to grow exponentially.

We derive the $L^2$ convergence rates with respect to  $\|\cdot\|_{F,2}$ when $x \sim F$ for a probability measure $F$ over $\mathcal{X}$, where
\begin{equation}
    \|r_g\|_{F,2} \equiv \sqrt{\int_{x\in\mathcal{X}} r^2_g(x)dF(x)},\quad \|r_g\|_{F,\infty} \equiv \sup_{x\in\mathcal{X}}|r_g(x)|,
\end{equation}
\noindent characterize the approximation properties of the underlying class of functions under $L^2(\mathcal{X},F)$ and uniform distances for any function $g\in \mathcal{G}$.

Our main result, Theorem \ref{theorem:l_2_rates}, extends the $L^2$ convergence result in \cite{belloni2015} by deriving new finite-sample rates as well as asymptotic rates using the results from \cite{rudelson2007sampling}. This result is valid for a large class of series regression models that satisfy assumption \ref{eq:assump_sample_model}. Therefore, it is of interest beyond the case of polynomial regression further discussed in this paper.

\begin{theorem}[$L^2$ rates]
\label{theorem:l_2_rates}
Consider the following assumptions
\begin{enumerate}
    \item For each $n$, random vectors $\{(y_i, x_i')\}_{i=1}^n$ are $i.i.d.$ and given by the series regression model (1) with $\bar{\sigma}^2 \equiv \sup_{x\in \mathcal{X}}E[\epsilon_i^2 |x_i=x] < \infty$.
    \item Uniformly over all $n$, eigenvalues of $Q\equiv E[p_ip_i']$ are bounded above and away from zero.
    \item For each $n$ and $k$ there exist a finite constant such that for all $g\in \mathcal{G}$
    $$
    \|r_g\|_{F,2} \leq c_k.
    $$
\end{enumerate}
 Then, if $c_k \to 0$
\begin{equation}
\|\hat g - g\|_{F,2}
= O_p\!\left[
\left(\xi_k \sqrt{\frac{\log n}{n}} + 1\right)
\left(\sqrt{\frac{k}{n}} + c_k\right)
\right],
\end{equation}

and if $\epsilon_i \sim \text{subG}(\sigma_i^2)$, for $t \in (0,1)$ such that $c_k \leq t/2$, and $\frac{c_k \xi_k}{\sqrt{n}}\lor\sqrt{k/n} < t/8$,
\begin{equation}
        P(\| \hat{g} - g\|_{F,2} > t)
    \lesssim \exp\left\{-\frac{t}{a^2}\right\} + \exp\left\{ -\frac{nt^2}{\xi_k c_k}\right\} + \exp\left\{ -\frac{n t^2}{\xi_k\bar{\sigma}^2}\right\},
\end{equation}
where $a = \xi_k\sqrt{\frac{\log n}{n}} $.
\end{theorem}

The main takeaways from the first part of Theorem \ref{theorem:l_2_rates} is that the error in estimating the target function $g$ by series regression is bounded by the sum of an estimation error term $\sqrt{\frac{k}{n}}$ and an approximation error term $c_k$. This bound highlights the dimensionality problem of polynomial regression: $\xi_k\lesssim k$ and $\sqrt{\frac{\log(n)}{n}}\xi_k \to 0$ together imply that $n$ has to grow at a rate of at least $k^2$. Since for a polynomial embedding $k = O(J^d)$, this quickly yields unfeasible sample sizes as $d$ increases. The second part of Theorem \ref{theorem:l_2_rates} provides finite-sample rates valid for all $n$ and $k$ when the bound on the approximation error is small enough. While this finite-sample rate is not of immediate practical interest as it does require large $n$ to be useful (and yield a rate smaller than one), it clarifies the role of the approximation and estimation errors. This situation is similar to the approximation rates developed in \cite{farrell2021deep} and \cite{yarotsky2017error} for neural networks with growing layer size, that also require large $n$ to be of practical relevance.

Theorem \ref{theorem:l_2_rates} can be used to characterize which models have the fastest rates for a given class of embeddings $p(x)$ and target functions spaces $\mathcal{G}$. Since different modeling settings imply different bounds for the terms $\xi_k$ and $c_k$, the finite-sample rate will depend on the modeling assumptions. In particular, in this paper we consider $\mathcal{G}$ to belong to $\Sigma_s(\mathcal{X})$, the class of functions of H\"older\ smoothness of order $s$ defined by

\begin{equation}
\Sigma_s(\mathcal X)
:= \Bigl\{ g:\mathcal X \to \mathbb R \; \Big|\;
\forall x,\tilde x \in \mathcal X,\;
|g(x)-g(\tilde x)|
\lesssim \Bigl(\sum_{j=1}^d (x_j-\tilde x_j)^2\Bigr)^{s/2}
\Bigr\}.
\end{equation}

when $s\in (0,1]$. The definition can be extended for $s>1$ by bounding the difference in derivatives (see the Online Appendix).

This implies that if $\mathcal{G}$ is contained in a ball of finite radius in $\Sigma_s(\mathcal{X})$, for the polynomial series
\begin{equation}
    c_k \lesssim k^{-s/d},
\end{equation}
which means the approximation error is bounded by the inverse of the number of terms in the polynomial series, see for example, \cite{newey1997convergence} for references. In this case, the approximation error becomes bounded by a function of order $J^{-s}$.

To solve the curse of dimensionality problem in series regression, we study \textit{partitioned} series regression models in which features are built group-wise.

\begin{definition}[Partitioned Polynomial Regression]
A series regression model in which features $(x_i^1, \dots, x_i^d)$ are divided in $B\in \{1,\dots, d\}$ equally sized groups and polynomial embeddings of order $J$ are generated independently for each group.
\end{definition}

A \textit{partitioned} series regression model with $B$ groups reduces the number of transformed features from $k = (J+1)^d-1$ to $k \leq B((J+1)^{\lceil d/B\rceil}-1)$. This change implies a trade-off: it decreases the estimation error by reducing the number of parameters to estimate, but it increases the approximation error. A rough bound on the approximation error for the \textit{partitioned} polynomial series regression model when $\mathcal{G}\subset \Sigma_s(\mathcal{X})$ is
\begin{equation}
    c_k \lesssim B^{-s/d}(J+1)^{-\lfloor s/B\rfloor},
\end{equation}
which is increasing in $B$. This bound, while not applying directly for BPR, is informative for ensemble polynomial regression models. Indeed, as shown by \cite{breiman1996bagging} bagged models with random sampling of features (like random forests), have lower generalization error than their aggregated counterparts. Hence, one can think of this bound as being useful for BPR in which $M$ polynomial models are built by generating polynomial features for $\lceil d/M\rceil$ randomly sampled features without replacement as then $M = B$. So, there is a \textit{trade off} for BPR between the approximation error and the estimation error that is parameterized by the number of weak estimators and sampled features per estimator. We study this trade-off directly in Section 3, by adapting Theorem \ref{theorem:l_2_rates} directly for BPR. Corollary \ref{corollary:opt_rate_bpr} relates the group size $B$ to the $L^2$ rate of Theorem \ref{theorem:l_2_rates}.

\begin{corollary}[Optimized rate for Partitioned Polynomial Regression]
\label{corollary:opt_rate_bpr}
Let $g\in \mathcal{G}$ where $\mathcal{G}$ is a bounded subset of $\Sigma_s(\mathcal{X})$. Under the assumptions of Theorem \ref{theorem:l_2_rates} when $k\asymp B(J)^{d/B}$ it follows that $c_k \lesssim B^{-s/d}( J)^{-\lfloor s/B\rfloor}$ and $\xi_k \lesssim B(J)^{d/B}$. Furthermore, for $t \in (0,1)$ such that $B^{-s/d}(J)^{-s/B} \leq t/2$, and large enough $n$, let $G$ be defined as follows
\begin{equation}
\small
    G(n,d,J,s,t, B) = \exp\left\{-\frac{nt}{\log(n)d^2(J/B)^{2d}}\right\} + \exp\left\{ -\frac{nt^2}{B^{(d-s)/d}(J/B)^{d-s}}\right\} + \exp\left\{ -\frac{n t^2}{d(J/B)^{d}}\right\},
\end{equation}
then,
\begin{equation}
        P(\| \hat{g} - g\|_{F,2} > t)
    \lesssim G(n,d,J,s,t, B),
\end{equation}
and given $n, d, J$, $s$ and $t$ there exists an optimized rate given by the function
\begin{equation}
        G^*(n,d,J,s, t) = \min_{B\in \mathbb{Z}^+_{\leq d}} G(n,d,J,s,t,B)
\end{equation}
that is achieved at
\begin{equation}
        B^* = \sup\{ B\in\{1,\dots, d\}|  B^{-s/d}(J)^{-s/B} \leq t/2\}
\end{equation}
when $d\geq s$.
\end{corollary}

Corollary \ref{corollary:opt_rate_bpr} states that when features are built at the rate of the leading term in the number of features for \textit{partitioned} polynomial regression or BPR, $k\asymp B(J)^{d/B}$, there exists an optimal finite-sample rate that minimizes the rate found in Theorem \ref{theorem:l_2_rates}. Furthermore, for a specific setting (fixing $n, d, J, s$ and $t$) the optimal group size $B^*$ is the largest possible $B$ that keeps the approximation error bound $c_k$ small enough with respect to $t$. The optimal $B^*$ is increasing in the degree $J$, the smoothness parameter $s$, and slowly decreasing in the feature space dimension $d$. Intuitively this result means that in settings in which approximating the target function is easier (low dimensions, smoother spaces, or being able to build polynomials of high degree) we are better off splitting the sample space in more groups as the estimation error will be more important than the approximation error.

The implications of Corollary \ref{corollary:opt_rate_bpr} are better seen in Figure 1. Panels (a) and (b) show that for smooth settings (high $s$) or when we have series of high polynomial degree (high $J$) it is optimal to partition the features in groups. In fact, in some very smooth cases it may be optimal to have groups of just 2 features (i.e. $B^* = d/2$). This relationship holds regardless of the tolerance parameter $t$ as shown by the confidence bands in the figure. Panel (c) shows that not partitioning the features in groups can lead to very slow rates as $d$ increases. Indeed when $B=1$ the rate rapidly increases with $d$, whereas when $B=10$ the rate is controlled even in large dimensions. This highlights the problem with standard polynomial regression and the potential benefits of using \textit{partitioned} or BPR in high dimensions. Finally, panel (d) shows that $B$ also has a big impact on the sample size required for the rate to converge to zero. In a smooth setting with $J=3$, only when $B$ is high do we have convergence when $n=10^8$. This again speaks to the slow rate of the standard polynomial regression and how \textit{partitioned} and how \textit{bagged} models may be able to offer a faster solution.

\begin{figure}[t]
\centering
\begin{minipage}{0.80\textwidth}
\centering
\begin{subfigure}[b]{0.49\linewidth}
  \centering
  \includegraphics[width=\linewidth]{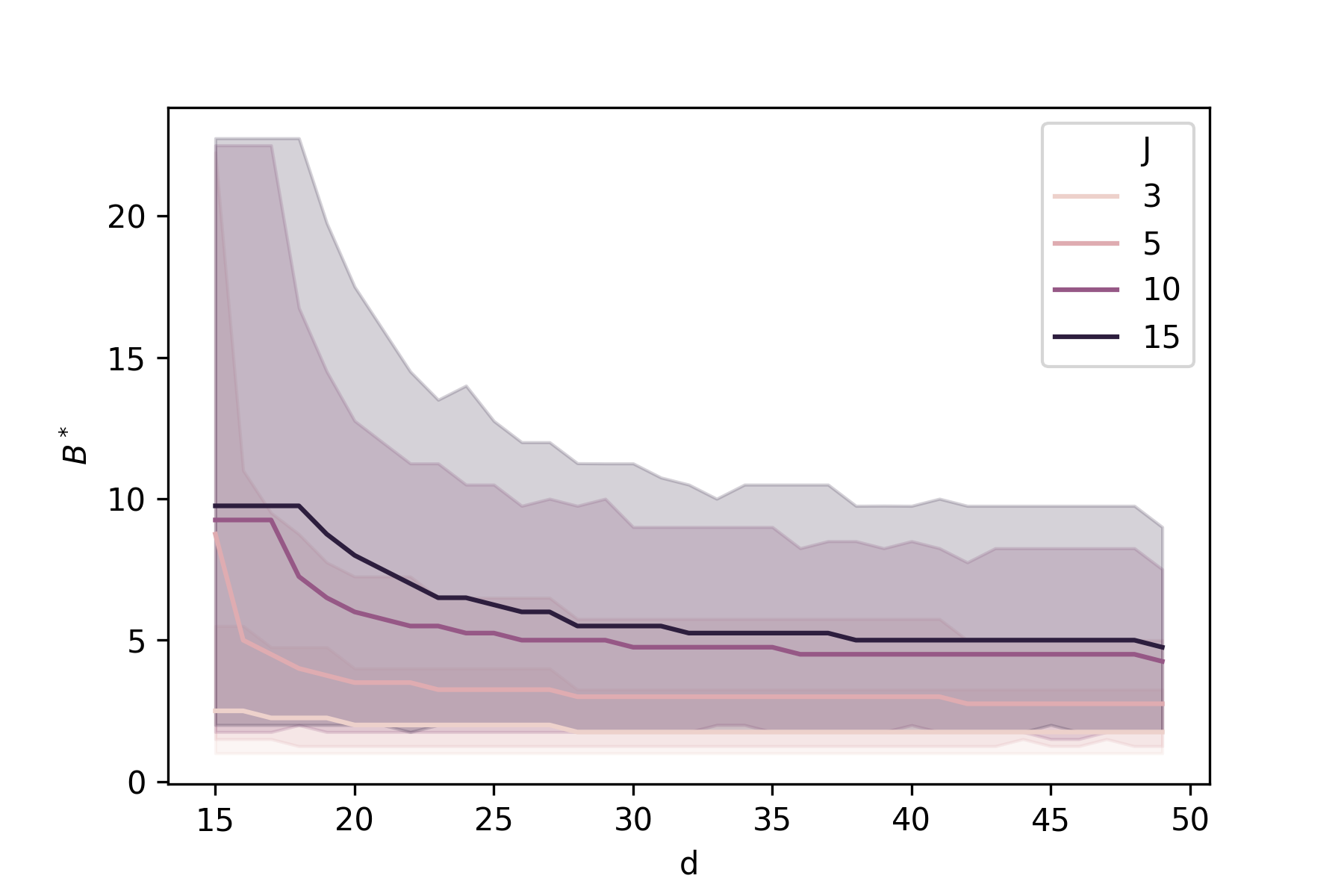}
  \caption*{(a) $B^*$ by $J$ and $d$.}
\end{subfigure}\hfill
\begin{subfigure}[b]{0.49\linewidth}
  \centering
  \includegraphics[width=\linewidth]{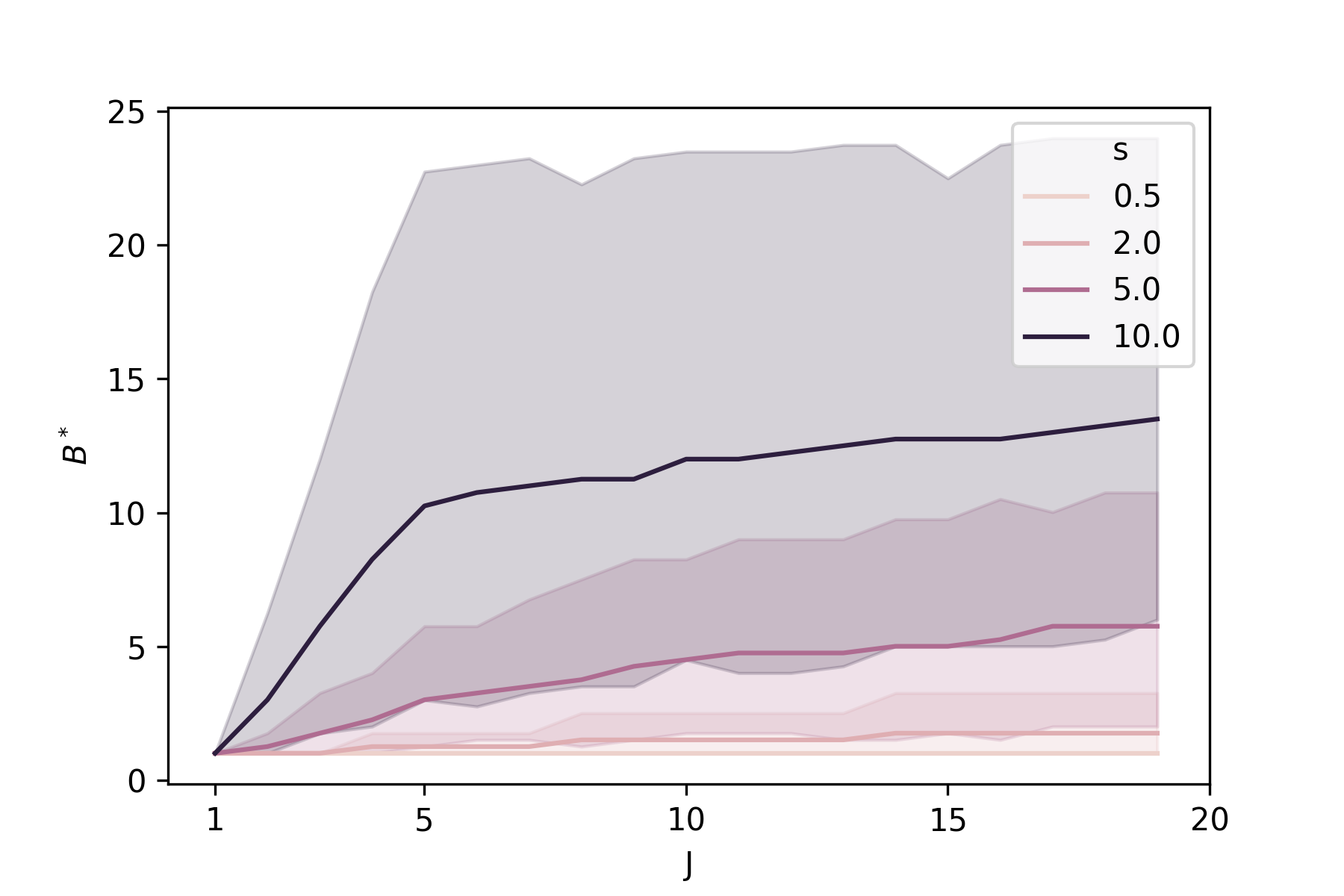}
  \caption*{(b) $B^*$ by $s$ and $J$.}
\end{subfigure}

\vspace{0.15em}

\begin{subfigure}[b]{0.49\linewidth}
  \centering
  \includegraphics[width=\linewidth]{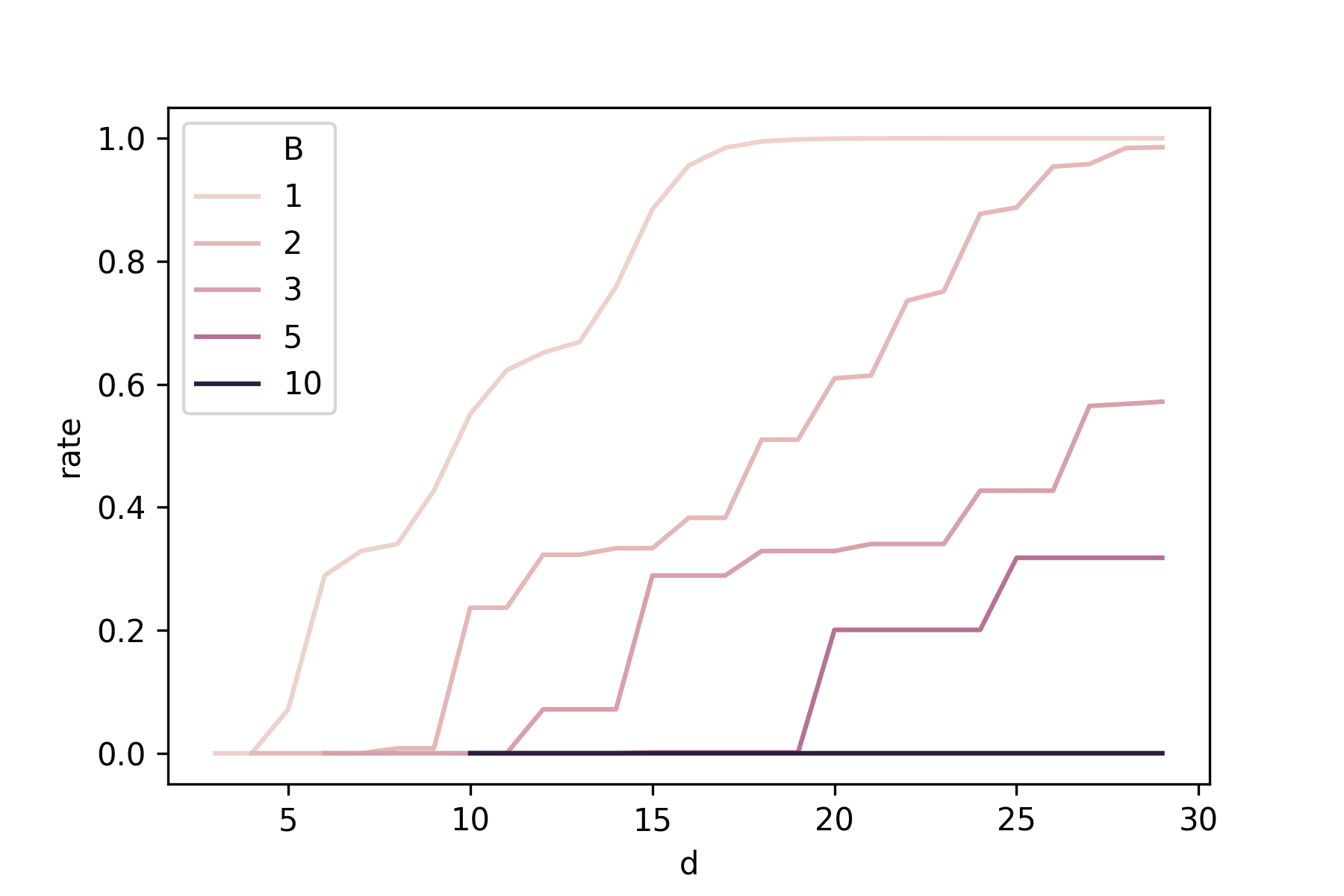}
  \caption*{(c) Rate by $B$ and $d$.}
\end{subfigure}\hfill
\begin{subfigure}[b]{0.49\linewidth}
  \centering
  \includegraphics[width=\linewidth]{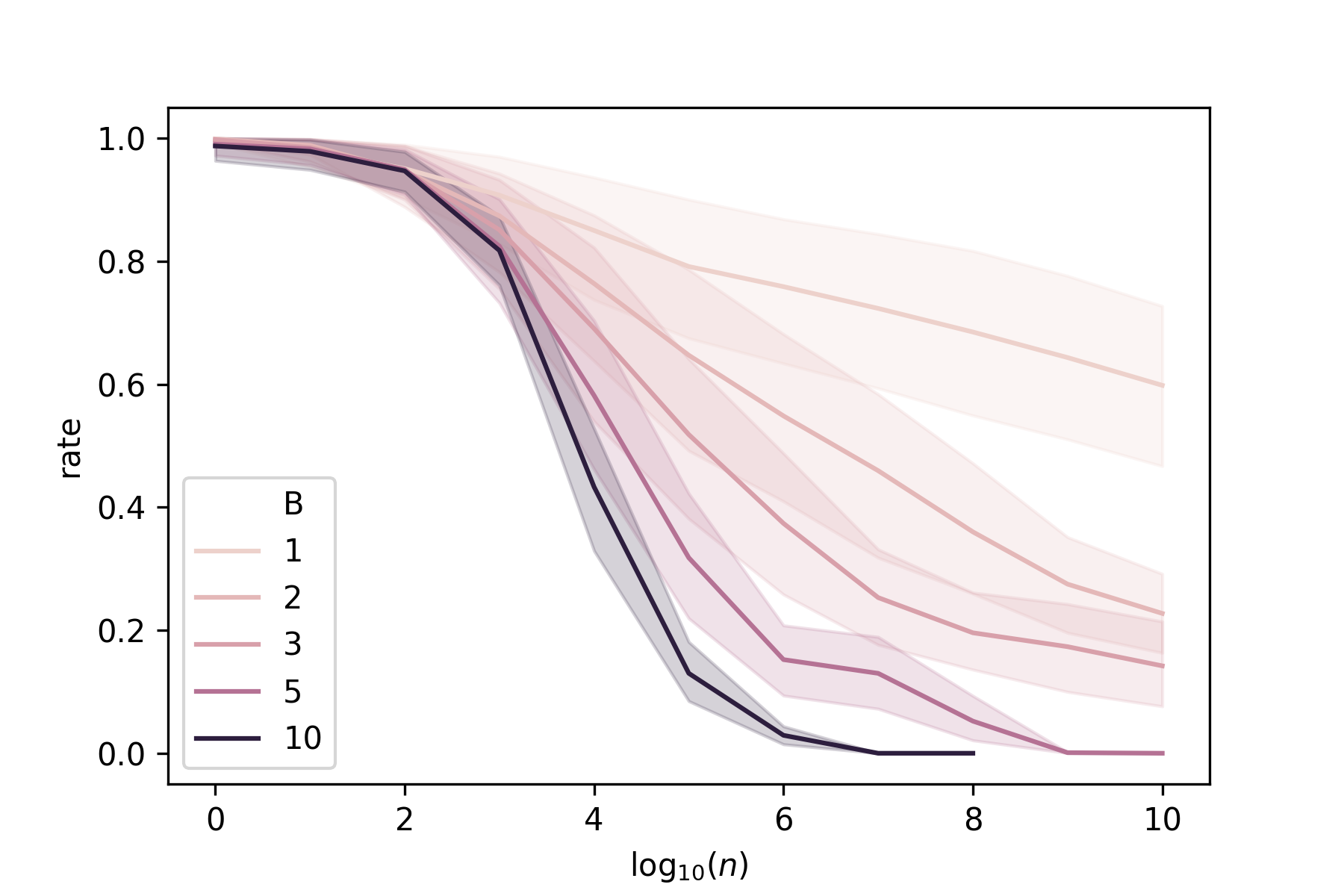}
  \caption*{(d) Rate by $B$ and $n$.}
\end{subfigure}

\end{minipage}

\vspace{0.1em}

\caption{\footnotesize Finite-sample rates and the choice of partition number $B$.}
\label{figure:finite_rate_partition_choice}

\vspace{-0.0em}
\begin{minipage}{0.95\linewidth}
\scriptsize
Notes: Panel (a) plots the upper bound on $B^*$ for different $J$ and $d$ when $s=5$. Panel (b) shows the upper bound on $B^*$ for different $s$ and $J$ when $d=40$. Solid lines/bands are means/confidence bands over $t\in[0.001,0.5]$. Panel (c) plots the finite-sample rate (normalized to be $\le 1$) over $B$ and $d$ for $J=3$, $n=10^7$, $s=5$, $t=0.05$. Panel (d) plots the same rate over $n$ with bands over $d\in[1,50]$. Jumps are due to $d/B$ being coarsened to integers.
\end{minipage}

\end{figure}

Beyond favorable estimation rates, researchers may also be interested in interpreting and understanding $\hat{g}$. A natural way of doing so is to inspect the marginal effects, or partial derivatives, of the estimated $g$ for particular features. Polynomial regression, and by extension PPR and BPR, offer a straightforward way of estimating derivatives. Since the polynomial basis is known, researchers can analytically compute the partial derivative given their estimate of $\hat{g}$ without requiring numerical approximations. For example, if the researcher is interested in the average marginal effect (or Average Partial Derivative) of feature $x_j$, 
$$
\text{APD}_j(\hat{g}) = \mathbb{E}\left[\frac{\partial\hat{g}(x)}{\partial x_j}\right],
$$
\noindent where the empirical expectation is taken over the observed sample. The researcher can directly compute $\frac{\partial\hat{g}(x)}{\partial x_j} = \nabla_jp(x)'\hat{\beta}$ by taking the derivative of the basis dictionary $p(x)$ with respect to feature $j$ for each data point $x$ and multiplying it by the estimated $\hat{\beta}$. Interestingly, the estimated marginals need not depend on all the elements of $\hat{\beta}$, but rather only on the elements for which $\nabla_j p(x)$ has non-zero entries. This observation directly translates into the $L^2$ estimation rates as Corollary \ref{theorem:ME_rates} shows by extending Theorem \ref{theorem:l_2_rates} to the estimation of marginal effects. 

\begin{corollary}[Marginal Effects $L^2$ rates]
\label{theorem:ME_rates}
In addition to assumptions 1-2 of Theorem \ref{theorem:l_2_rates} consider
\begin{enumerate}
    \item The derivative of $p(x)$ with respect to $x_j$ is supported at most on $s_j < k$ points. That is for all $j \in \{1,\dots, d\}$, $|\mathcal{S}_j| \leq s_j$ for 
    $$
    \mathcal{S}_j = \{ i \in \{1,\dots, k\} \mid [\nabla_j p(x)]_i \neq 0\}.
    $$
    \item For all $j \in \{1,\dots, d\}$, there exists a finite constant $C_{\partial} > 0$, independent of $n$ and $k$, such that for $Q^{(j)} \equiv E[(\nabla_j p_i) (\nabla_j p_i)']$,
$$
\|Q^{(j)}\|_{op} \leq C_{\partial}.
$$
    \item For each $n$ and $k$ there exist a finite constant such that for all $g\in \mathcal{G}$ and for all $j\in \{1,\dots,d\}$
    $$
    \|\partial_j r_g\|_{F,2} \leq c'_k.
    $$
    Then, if $c'_k \to 0$ 

\begin{equation}
        \| \partial_j\hat{g} - \partial_j g\|_{F,2} = O_p\left(\left(\sqrt{\frac{\log(n)}{n}}\xi_k + 1\right)\left(\sqrt{\frac{s_j}{n}} + c_k\right)\right).
\end{equation}
\end{enumerate}
\end{corollary}

Estimating the partial derivatives of $g$ further highlights the trade-offs between function approximation and parameter estimation. Compared to Theorem \ref{theorem:l_2_rates}, Corollary \ref{theorem:ME_rates} shows that the sampling error scales with $\sqrt{\frac{s_j}{n}} \leq \sqrt{\frac{k}{n}}$, where $s_j$ is the support of $\nabla_j p(x)$. Intuitively, the derivative estimate for feature $j$ only uses $s_j$ of the estimated $\hat{\beta}$s, so the estimation error due to the sampling variance scales with $s_j$ rather than $k$. Note that because the estimation of $\hat{\beta}$ still requires using all parameters $k$, the basis approximation component $\sqrt{\frac{\log(n)}{n}}\xi_k$ remains the same as in Theorem \ref{theorem:l_2_rates}. 

Furthermore, the improvement in the rate comes at the cost of a larger approximation error. Corollary \ref{theorem:ME_rates} assumes that $c'_k$, the approximation error for the average derivative over $x$, is small.\footnote{Observe that in Corollary \ref{theorem:ME_rates} the rate in (2.18) includes $c_k$ and not $c_k'$, this is because $c_k'$ is assumed to go to zero and so is not added to the expression, and the $c_k$ term appears because the proof uses a bound on the estimation $L_2$ error of $\hat{\beta}$ as in Theorem \ref{theorem:l_2_rates}, so the estimation error also depends on $c_k$.} Naturally, this is a stronger requirement and necessitates more degrees of smoothness than we require for the approximation of function $g$. For example, for the case in which $g\in\Sigma_s(\mathcal{X})$ and the basis dictionary is the polynomial series we have that
$$
c_k \leq c'_k \lesssim k^{-(s-1)/d}.
$$

\noindent Overall, however, if we believe we are in a setting with small approximation errors, as we posit in this paper, and $c'_k \to 0$, Corollary \ref{theorem:ME_rates} highlights an interesting point. Among models with similar approximation error, if we are interested in interpreting model estimates by computing marginal effects or APDs, models with simple known basis dictionaries $p(x)$, as used for PPR or BPR, have both easier to compute marginals and faster estimation error rates than more complex models. Therefore, one might say, that conditional on the approximation error, simpler models like PPR or BPR are more interpretable as they provide a more reliable way to estimate how the model changes with its inputs. This is a key point that we explore further in the empirical application in which simpler BPR models, with a smaller number of interactions, have simpler and more theoretically consistent APDs than more complex NNs with the same out-of-sample accuracy.

\section{Bagged Polynomial Regression with Random Projections}

Motivated by the theoretical insights we propose a \textit{bagged} polynomial regression method with random projections. The aim of the method is to address the computational problem and slow rates when we have to estimate a large number of parameters while maintaining interpretability. We do so by relying on the feature splitting idea of \textit{partitioned} series regression (PPR) by constructing models on random projections of the features, through explicit regularization via $\ell_2$ penalization, and by model ensembling through bagging. This method draws from the drop-out idea for neural networks of only using some neurons in each training iteration \cite{srivastava2014dropout}, the random projections literature (see \cite{iterForestsYu} for an iterative random forests estimator), and from the model ensembling and bagging literature \cite{breiman2001random}.

The BPR averages $M$ regression models trained on polynomial embeddings of degree $J$ of a random subset of basic features of size $F$. This setup allows us to choose $F$ and $J$, to determine the overall number of features for each model, which we denote by $K$, and control model complexity, making the problem computationally feasible even when we build polynomials of high degree. More precisely, we can write the output of the BPR model as
\begin{equation}
    \hat{y}_i = \frac{1}{M}\sum_{m=1}^M \hat{g}_m(x_i),
\end{equation}
\noindent where $\hat{g}_m(x_i) = \langle \hat{\beta}^m, \Psi^J(x_i^{F_m})\rangle$ are based models trained on an $N_m$ sample of observations of size $|N_m|$, with $\hat{\beta}^m$ being the weights for model $m$, and $\Psi^J$ is a function that generates polynomial features of degree $J$ from a (uniformly) random set of features denoted by $X_i^{F_m}$ of size $F$. The polynomial regression estimator for a model $m$ is defined by the $\hat{W}^m$ weights computed by solving the least squares problem for the restricted sample $N_m$ of observations and the polynomial features generated by $\Psi^J(X_i^{F_m})$. For simplicity, we assume that $S = |N_m|$ for all $m$, so that all models are trained on random samples of the same size. We include an $\ell_2$ regularization term denoted by $\lambda$, to avoid collinearity problems and improve out-of-sample fit, that can be hyper-tuned, along with the other parameters, using cross-validation. The following algorithm details the procedure:

\begin{algorithm}[H]
\footnotesize
\SetAlgoLined
\KwOut{$\hat Y$.}
\KwIn{$(Y,X)$; degree $J$; features $F$; bags $M$; subsample $S$; ridge $\lambda$.}

\For{$m=1,\dots,M$}{
  $F_m \leftarrow$ sample $F$ features\;
  $N_m \leftarrow$ sample $S$ indices\;
  form $\Psi^{J}(x_i^{F_m})$\;
  $\hat{\beta}^{m} \leftarrow \arg\min_{\beta}\ \sum_{i\in N_m}\!\bigl(Y_i-\langle \beta,\Psi^{J}(x_i^{F_m})\rangle\bigr)^2 + \lambda\|\beta\|_2^2$\;
  $\hat Y_i^{m} \leftarrow \langle \hat{\beta}^{m}, \Psi^{J}(x_i^{F_m}) \rangle$\;
}
$\hat Y_i \leftarrow \frac{1}{M}\sum_{m=1}^{M}\hat Y_i^{m}$\;
\caption{Bagged Polynomial Regression with Random Projections}
\label{algorithm:BPR}
\end{algorithm}

\noindent By hyper-tuning the parameters $J$, $F$, $M$, $S$ and $\lambda$, BPR optimally chooses how much to partition the feature space to build the polynomial features to navigate the approximation vs. estimation error trade-off. Choosing the hyper-parameters through cross-validation can be thought as estimating the optimal $B^*$ in Corollary \ref{corollary:opt_rate_bpr} given the data setting. Furthermore, the additional $\ell_2$ penalty and model ensembling ensure that the generalization error is minimized. Theorem 1.1 in the Online Appendix Section S1. extends Theorem \ref{theorem:l_2_rates} to bagging random projection estimators such as BPR under equivalent assumptions. The results highlight a similar trade-off, with the $L^2$ rate as $\bar{c}_K \to 0$,
\begin{equation}
\label{eq:l2_rate_bpr}
\|\hat g^{\mathrm{BPR}}-g\|_{F,2}
=
O_p\!\left(\left(\sqrt{\frac{\log S}{S}}\xi_K+1\right)\left(\sqrt{\frac{K}{S}}+\bar{c}_K\right)\right),
\end{equation}
\noindent where $\bar{c}^2_K \equiv E_K[\|g  - \Pi_Kg\|^2_{F,2}]$ bounds the average approximation error of the random projections of dimension $K$ over the distribution of possible $K$ dimensional subset of features and $\xi_K\equiv \sup_{x\in\mathcal X}\ \sup_{|F|=K}\|p_F(x)\|$. The bound differs from that of Theorem \ref{theorem:l_2_rates} in three important ways. First, the sampling estimation error scales with $\sqrt{\frac{K}{S}}$ rather than $\sqrt{\frac{k}{n}}$. This can be favorable in cases in which $k$ is much larger than $K$ even when we only use a small subsample for each base model ($S<n$). Second, the $\sqrt{\frac{\log S}{S}}\xi_K$ term requires the number of subsamples used to grow slower than $n$ is required to grow by Theorem \ref{theorem:l_2_rates} as $\xi_K < \xi_k$. Intuitively, this is because, contrary to the results for PPR, each base model need not estimate all $k$ parameters generated on all features and so the polynomial basis term in the bound only depends on $K$ parameters. Third, this comes at the cost of a larger approximation error that is bounded by $\bar{c}^2_K$ rather than $c_k$, which in general cases could be large. 

We explore conditions under which $\bar{c}^2_K$ is of the order of $c_k$ in Theorem 1.2 in the Online Appendix. Sufficient conditions for $\bar{c}^2_K$ to be well behaved, under uniform feature sampling, are that $K \to k$ or that no one feature is uniquely important to approximate the true function $g$ (that is, $\|\beta_g\|^2$ is small). In such cases, $\bar{c}^2_K \asymp c_k$ and the approximation error for BPR goes to zero as $c_k \to 0$. While these conditions might seem restrictive, in this paper we are interested in cases in which researchers are confident they can approximate well the true function $g$ and simultaneously want to interpret their model by inspecting relevant marginal effects. In such cases, even if  $\|\beta_g\|^2$ is small and so no one feature uniquely matters for estimating $g$, the researcher might still want to understand how their model varies along a particular feature. For example, in many economic settings, estimated models have moderate or low $R^2$, but that is not detrimental to the research question as researchers are interested on the model behavior for a particular parameter. 

Nevertheless, it is possible to improve the theoretical results for BPR with random projections and related estimators by modifying the algorithm to allow for non-random sampling of features (e.g. through IPW) or by using iterative procedures with variable selection steps. We refer interested readers to \cite{iterForestsYu} for estimators with these features in the case of random forests. On the other hand, the theoretical results for BPR can be extended to the case of marginal effect estimation studied in Theorem \ref{theorem:ME_rates}, when APDs are estimated by averaging the estimated derivatives for each base model. In the following section we detail how to do so in the context of our empirical application and note that the same trade-offs highlighted by Theorem \ref{theorem:ME_rates} also extend to BPR.

Finally, BPR with random projections can be used for both classification and regression problems. In the case of classification problems, we use logistic regression and minimize the logistic (cross-entropy) loss in step 5 of Algorithm 1, and instead of taking the average of the $M$ models we take the majority vote (median). In this case, bagging can be particularly useful as the activation function is non-linear so the models we train will have low bias and high variance. As in random forests, by taking the median/average of $M$ models we control the variance of the overall model and achieve a lower generalization error.

\section{Application: Crop classification from satellite imagery}

\subsection{Setting and data}

In this section, we apply BPR to satellite-based crop classification, a natural testbed because accurate crop maps are a core input to environmental and agricultural policy—supporting crop inventories, crop insurance administration, yield estimation, and enforcement of planting and quota regulations \parencite{zhu2017improving}.

We use data provided by \cite{khosravi2019random}. 
The crop images were collected by RapidEye satellites (optical) and the Unmanned Aerial Vehicle Synthetic Aperture Radar (UAVSAR) system (Radar) over an agricultural region near Winnipeg, Manitoba, Canada on 2012. Seven crops are included in this dataset: corn, peas, canola, soybeans, oats, wheat, and broadleaf. We approach the prediction task as a classification problem with seven classes. The data has 325,834 observations and 174 features.

\subsection{Predictive performance}

The model for predicting the seven crops combines seven crop-specific BPR classifiers (one per crop). We find that BPR reaches great performance even with a small ensemble as long as we have a larger feature budget: with 10 bags, 50 features, degree-2 polynomial embeddings, we achieve a test accuracy of 0.997. This is not only strong in absolute terms, but also competitive with (and slightly above) the best neural-network baselines commonly reported for this dataset (around 0.992), as summarized in publicly available benchmarks.\footnote{See alternative models trained on the same data at \url{https://www.kaggle.com/code/gandhoogle/winnipeg-mapping}.} Notably, these gains come without increasing model complexity beyond a low-degree (degree 2) embedding and a modest number of bags, highlighting that BPR can match top-performing black-box methods while retaining a simple, structured specification. Additional BPR specifications and their corresponding test accuracies are reported in the Online Appendix. In an additional application, our estimator also performs close to state-of-the-art prediction methods in the benchmark MNIST handwritten digit dataset, results are reported in the Online Appendix.

\subsection{Interpreting the fitted classifier}

In environmental economic applications, prediction accuracy is not the only goal: researchers and policy users often want diagnostics that clarify \emph{which inputs} the classifier is using and whether its learned associations align with basic domain expectations. This is especially relevant in remote sensing, where feature sets are high-dimensional and strongly collinear, so flexible models may rely on distributed representations that are difficult to audit.

BPR is well-suited to this setting because it is an ensemble of regularized polynomial generalized linear models. As a result, interpretability diagnostics are available by construction: we can (i) inspect coefficients within base learners and summarize them across the ensemble, and (ii) compute partial dependence and ICE curves for canonical remote-sensing covariates.

The covariate we use as an example in this section is NDVI, a standard vegetation index that proxies for canopy ``greenness'' and biomass. Holding the observation date fixed (early July), soybeans are typically at an earlier growth stage than many competing crops in this region, implying less canopy closure and therefore lower NDVI \parencite{ManitobaCropReport2025_07_08}. Thus, in a one-vs-rest soybean classifier, we expect higher NDVI (on July 5) to \emph{decrease} the predicted probability of soybeans, all else equal.

\paragraph{Coefficient-averaging diagnostic.}
To obtain a transparent summary of how a specific covariate is used across the ensemble, we compute a simple coefficient-averaging diagnostic. Fix a covariate $X_j$ (here, NDVI on July 5). BPR fits many polynomial logistic regressions on randomly selected feature subsets; some base learners include $X_j$, others do not. We extract the coefficient on the \emph{linear} term in $X_j$ from each base learner that includes $X_j$ and report its mean and standard error across those learners. This simple summary is an estimate of the Marginal Effect of $X_j$ when the other covariates are fixed at zero ($X_{-j} = 0$) which we denote by $\text{ME}_j(0)$. Furthermore, because our logit index is a polynomial in the covariates, we can easily compute and report the Average Partial Derivative (APD) of NDVI as discussed in the theoretical discussion of Corollary \ref{theorem:ME_rates},

\begin{align}
\hat g(x)
&\;\equiv\; \frac{1}{M}\sum_{m=1}^M \hat g_m(x),
\qquad
\hat g_m(x)\;\equiv\;\Lambda\!\left(\eta_m(x)\right),
\qquad
\eta_m(x)\;\equiv\;\alpha_m+\hat\beta_m^\top \Psi_J(x^{F_m}),
\label{eq:bpr_pred_def}\\[4pt]
\operatorname{APD}_j
&\;\equiv\;
\mathbb{E}\!\left[\frac{\partial}{\partial x_j}\hat g(X)\right]
\;=\;
\frac{1}{M}\sum_{m=1}^M
\mathbb{E}\!\left[
\Lambda\!\left(\eta_m(X)\right)\Big(1-\Lambda\!\left(\eta_m(X)\right)\Big)\cdot
\frac{\partial \eta_m(X)}{\partial x_j}
\right],
\label{eq:apd_bpr}
\end{align}
where $\Lambda(u)\equiv (1+e^{-u})^{-1}$.

\begin{table}[!htbp]
\centering
\caption{Coefficient-averaging diagnostic for NDVI (soybeans vs rest)}
\label{tab:soy_ndvi_coef}
\renewcommand{\arraystretch}{1.15}
\setlength{\tabcolsep}{7pt}
\begin{tabular}{lcrr}
\hline
Outcome & Type &  Mean coef & SE \\
\hline

Soybeans vs rest & ME(0) &  -0.121 & 0.051 \\
Soybeans vs rest & APD &  -0.575 & 0.161 \\
\hline
\end{tabular}
\begin{minipage}{0.9\linewidth}
\footnotesize
\emph{Note:} Each row summarizes effect of the NDVI variable across base learners (bags) that include NDVI\_Opt05July. BPR models were run with 80 bags, 50 features, and degree-2 polynomial embeddings.
\end{minipage}
\end{table}
\vspace{-1.6\baselineskip} 
\paragraph{PDP and ICE diagnostics.} As a second, more functional diagnostic, we study how predicted class probabilities vary with a focal covariate using partial dependence plots (PDP) and individual conditional expectation (ICE) curves, standard diagnostic measures for NNs. For a fitted classifier $\hat g(x)$ and covariate $X_j = z$, the partial dependence function is
$
\mathrm{PDP}(z) = \mathbb{E}\big[\hat g(z, X_{-j})\big],
$
where the expectation is taken over the empirical distribution of $X_{-j}$; ICE curves compute $\hat f(z, x_{-j,i})$ for each observation $i$. We report PDPs with ICE overlays to visualize both average response and heterogeneity.

\begin{figure}[htbp]
\centering
\begin{minipage}{0.60\textwidth} 
\centering

\begin{subfigure}[b]{0.49\linewidth}
  \centering
  \includegraphics[width=\linewidth]{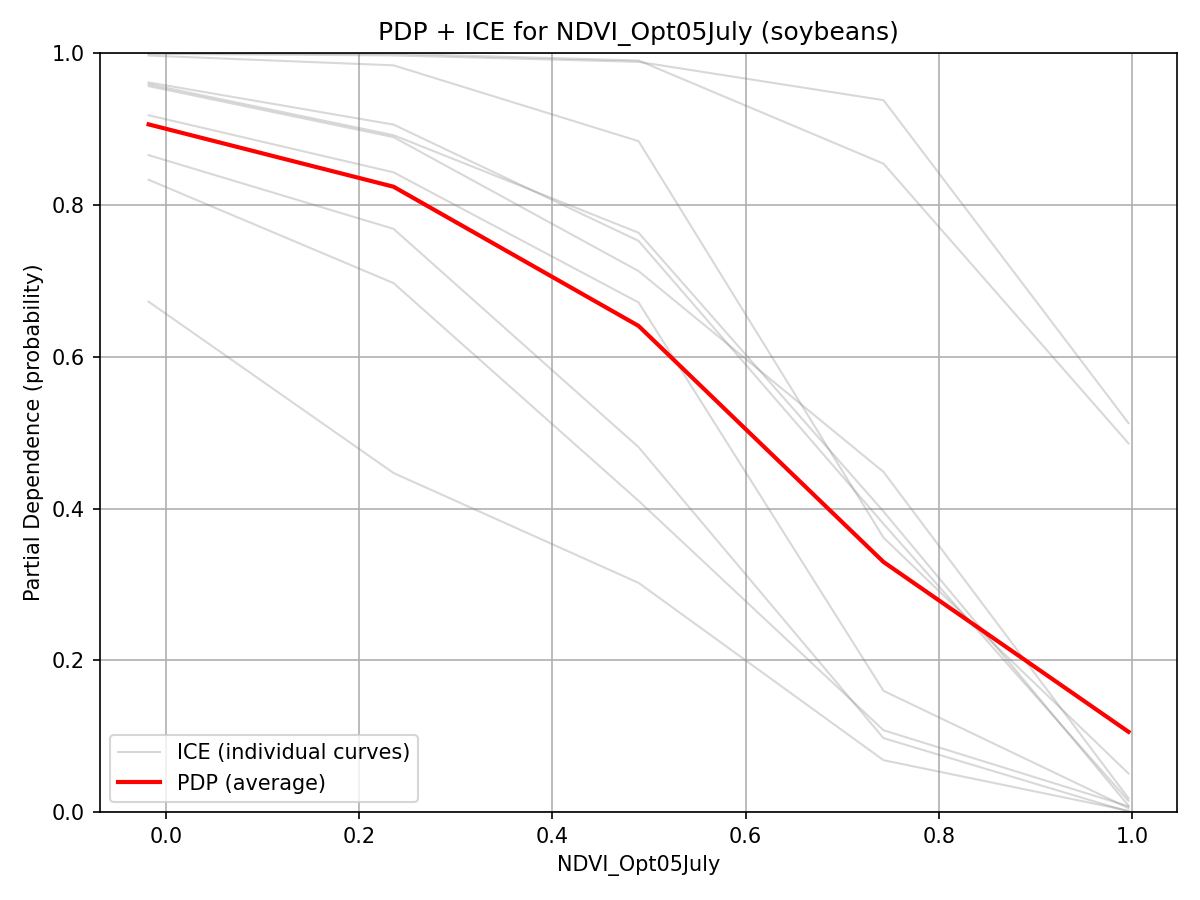}
  \caption{BPR (Soybeans)}
\end{subfigure}\hfill
\begin{subfigure}[b]{0.49\linewidth}
  \centering
  \includegraphics[width=\linewidth]{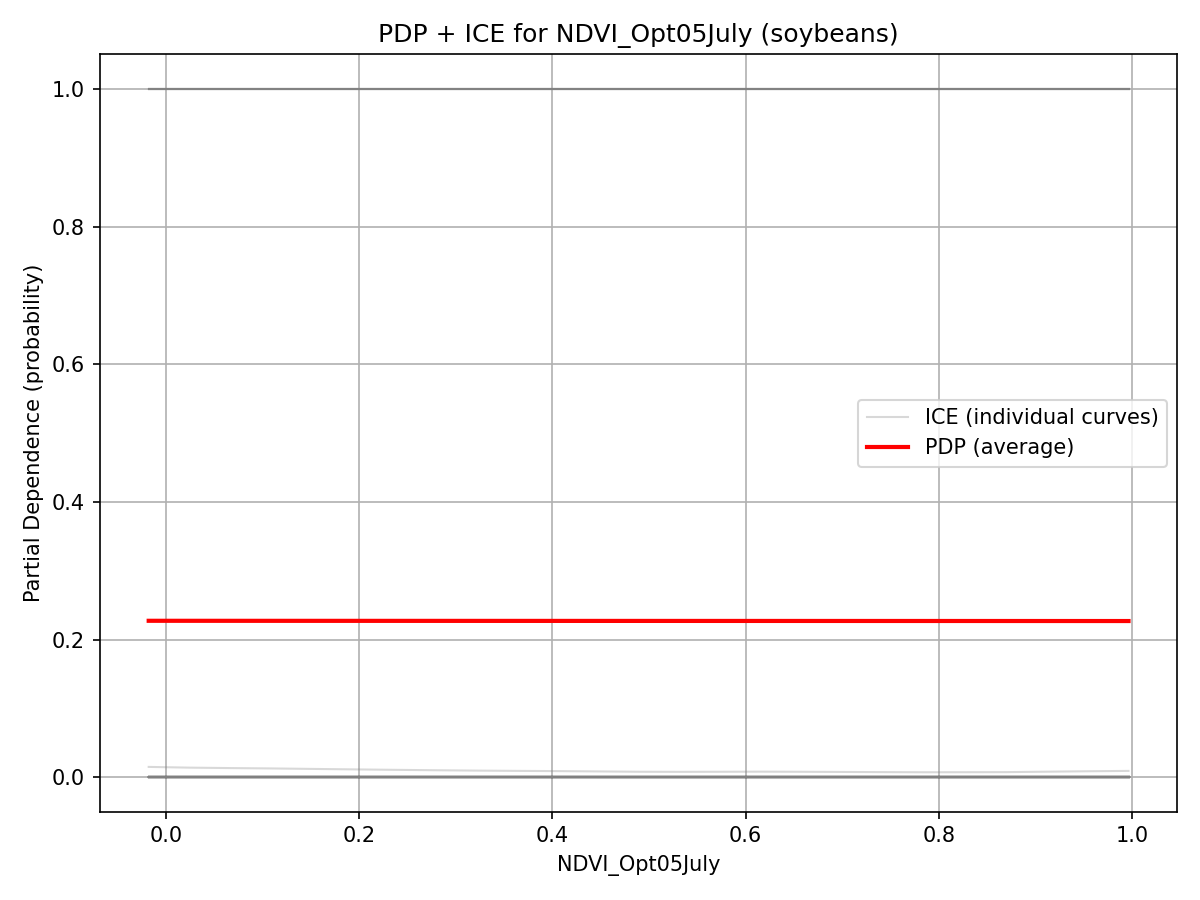}
  \caption{NN (Soybeans)}
\end{subfigure}

\vspace{0.25em}

\begin{subfigure}[b]{0.49\linewidth}
  \centering
  \includegraphics[width=\linewidth]{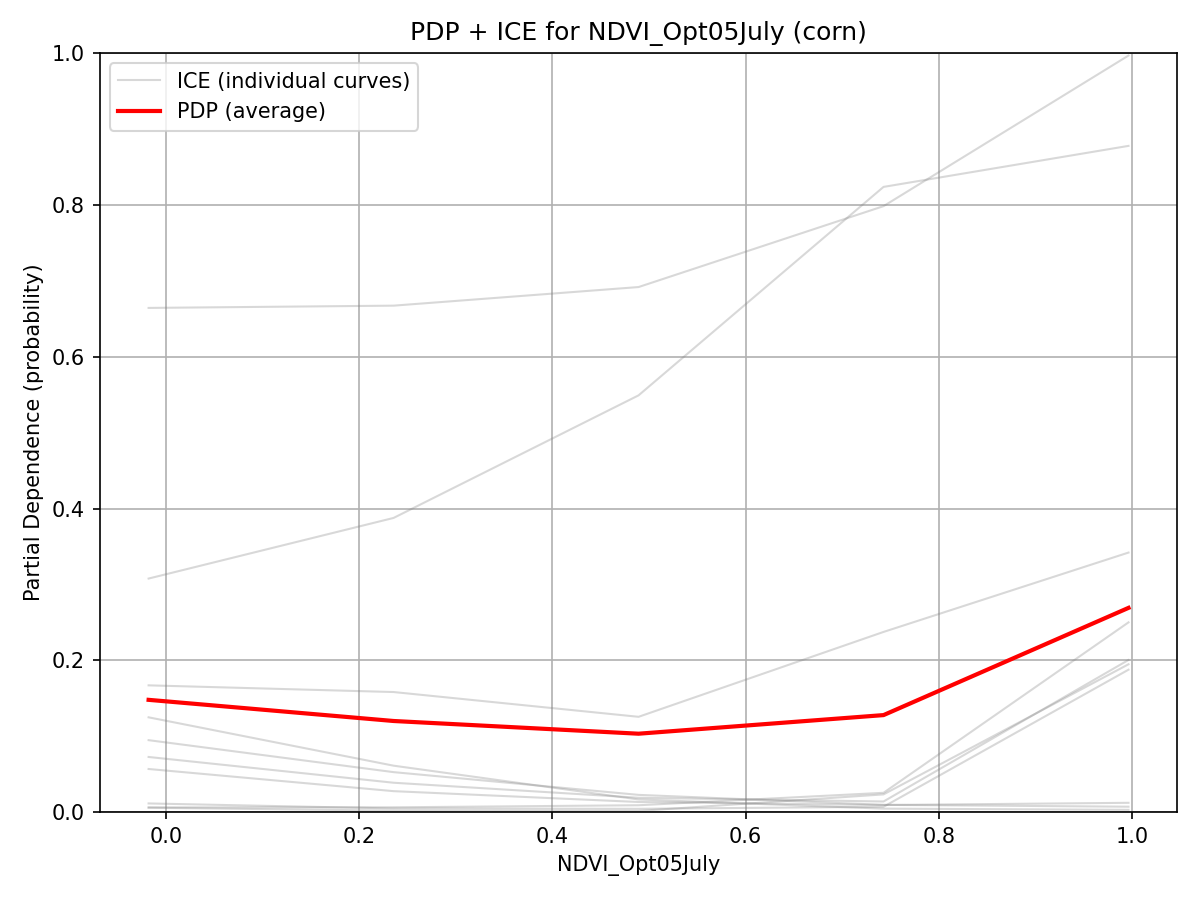}
  \caption{BPR (Corn)}
\end{subfigure}\hfill
\begin{subfigure}[b]{0.49\linewidth}
  \centering
  \includegraphics[width=\linewidth]{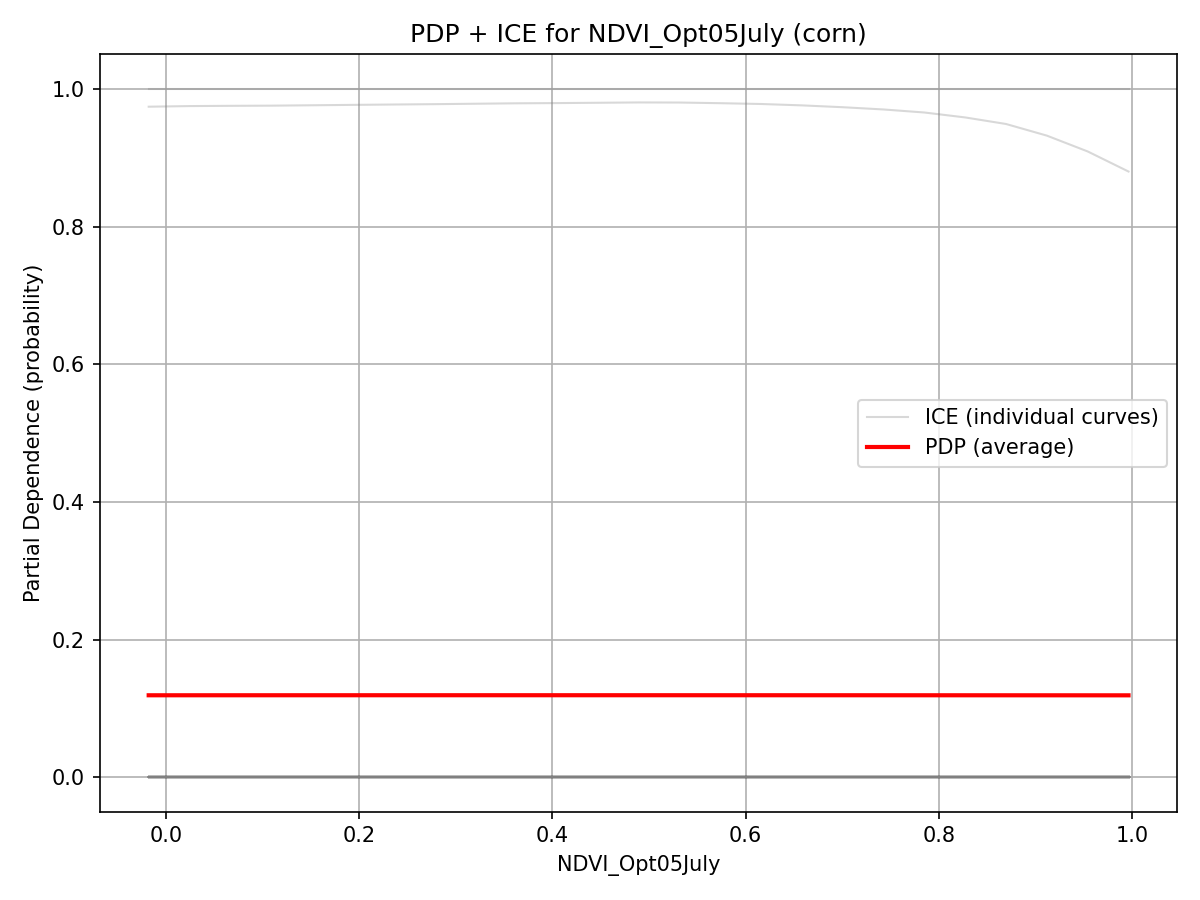}
  \caption{NN (Corn)}
\end{subfigure}
\end{minipage}
\caption{\scriptsize Partial dependence of the fitted classifier on NDVI\_Opt05July. Panels compare BPR and a neural network for soybeans (top row) and corn (bottom row). For BPR, curves are computed using only base learners that include NDVI\_Opt05July. BPR models were run with 50 bags, 10 features, and degree-2 polynomial embeddings. NN models were run with PyTorch and 2 hidden layers (128/64), dropout 0.3, weight decay 1e-4, 40 epochs. 
}
\label{fig:pdp_ndvi_05july_bpr_vs_nn_soy_corn}
\end{figure}

Figure \ref{fig:pdp_ndvi_05july_bpr_vs_nn_soy_corn} reports PDP/ICE diagnostics for NDVI on July 5th for soybeans (top row) and corn (bottom row), comparing BPR to a neural network. Consistent with agronomic intuition for early July, the BPR soybean classifier exhibits a clear negative dependence on NDVI (Panel (a)): higher NDVI is associated with a lower predicted probability of soybeans relative to other crops.  In contrast, the BPR corn classifier shows a much weaker and less systematic relationship (Panel (c)), consistent with NDVI being less diagnostic for corn at this date in our sample. The intuition for the PDP plots is matched by our coefficient summary measures in Table \ref{tab:soy_ndvi_coef}. Both the APD and the ME at zero indicate that increasing NDVI decreases the probability of a crop being classified as soybeans. We find that on average increasing NDVI from 0 to 1 is associated with a -.575 average change in the predicted soybean probability (APD). In the Online Appendix, we provide the PDP plots for BPR for five additional crop types and find similarly intuitive patterns. 

A striking contrast is that the neural network PDPs are essentially flat for both crops (Panels (b) and (d)), despite the neural network achieving test accuracy comparable to BPR. This pattern is consistent with a well-known limitation of marginal dependence diagnostics for NN: with many related remote-sensing predictors, a flexible neural network can encode the relevant signal in distributed representations across correlated substitutes, so variation in a single raw covariate $X_j$ need not move $\hat g(z, X_{-j})$ holding $X_{-j}$ fixed. As a result, the functional $\mathbb{E}[\hat g(z, X_{-j})]$ can be nearly invariant in $z$ even when $X_j$ is predictive, leading to a misleading estimate of the true marginal effect. By contrast, BPR’s randomized feature sub-sampling implies that, conditional on $X_j$ being included, some base learners must directly load on $X_j$ in a simpler analytical model, yielding PDP/ICE curves that more transparently reflect how the fitted classifier uses NDVI.\footnote{PDP/ICE plots for all crops and are reported in Online Appendix.}

\section{Conclusion}

In this paper, we highlight why polynomial regression, while offering more interpretability than neural networks and being able to approximate the same function classes, is rarely used in practice. By deriving new finite-sample and asymptotic $L^2$ rates for series regression estimators, we show that the convergence rate for polynomial regression can be very slow. However, the rate can be improved when the polynomial embeddings are generated group-wise for partitions of the feature space rather than for all features. The improvement is particularly salient when the function class we are trying to estimate is smooth and for the estimation of marginal effects. Motivated by these results, we propose the use of bagged polynomial regression with random projections instead of standard polynomial regression as a potential substitute for neural networks in high-dimensional applications. BPR draws from the theoretical insight of building polynomial embeddings for subsets of the feature space and from ensembling models by averaging multiple estimators to improve the out-of-sample performance. For an environmental economic prediction problem we show that it can perform similarly to more complex NNs, and deliver better estimates of marginal effects, leading to superior model interpretability that matches scientific priors.


\bibliography{references}


    \section*{Appendix A: Proofs of Results}
    \renewcommand{\theequation}{A.\arabic{equation}}
    \renewcommand{\thesection}{A}
    \setcounter{equation}{0}

    \medskip

\appendix
\setcounter{section}{0}
\setcounter{equation}{0}
\setcounter{theorem}{0}
\setcounter{corollary}{0}
\renewcommand{\thesection}{A.\arabic{section}}
\renewcommand{\theequation}{A.\arabic{equation}}
\renewcommand{\thetheorem}{A.\arabic{theorem}}
\renewcommand{\thecorollary}{A.\arabic{corollary}}

\section{Proof of main results}
\subsection{Proof of Theorem \ref{theorem:l_2_rates}}
\begin{proof}
This proof adapts \cite{belloni2015} by using \cite{rudelson2007sampling} (details in the Online Appendix) instead of \cite{rudelson1999random} and deriving the asymptotic rates for non leading terms. Then the results are extended to include a finite-sample rate under a subgaussianity assumption. Notation wise let $\| \cdot \|$ denote the Euclidean norm and $\| \cdot \|_{op}$ denote the operator norm for matrices. It will be useful to note that for a matrix $A$ and vector $b$, $\| A b\| \leq \|A\|_{op} \|b\|$.

Start by observing that
\begin{equation}
    \| \hat{g} - g\|_{F,2} \leq \|p'\hat{\beta} - p'\beta\|_{F,2} + c_k
\end{equation}
and under normalization $Q=I$
\begin{equation}
    \|p'\hat{\beta} - p'\beta\|_{F,2} = \|\hat{\beta} - \beta\|.
\end{equation}
By the triangle inequality,
\begin{equation}
    \| \hat{\beta} - \beta\| \leq \|\hat{Q}^{-1} \mathbb{E}_n[p_i\epsilon_i]\| + \|\hat{Q}^{-1} \mathbb{E}_n[p_ir_i]\|.
\end{equation}
\noindent where $\hat{Q} = \frac{1}{n}\sum_i p_i \otimes p_i$.
Applying \cite{rudelson2007sampling} for $p_i$ we get that
\begin{equation}
    E\|\hat{Q} - Q\| \leq a,
\end{equation}
when $a\leq 1$ for $a = C \sqrt{\frac{\log n}{n}} \xi_k$ for a finite constant $C$. Furthermore, it can be shown that all eigenvalues of $\hat{Q}$ are bounded away from zero so the inverse exists and is well defined. This allows us to apply the continuous mapping theorem for the inverse. Next, we bound the two terms that depend on $\hat{Q}^{-1}$.

By Markov's inequality, for $t>0$
\begin{align}
    P(\|\hat{Q}^{-1} \mathbb{E}_n[p_i\epsilon_i]\| >t) &\leq \frac{E\|(\hat{Q}^{-1} - Q + Q) \mathbb{E}_n[p_i\epsilon_i]\|}{t} \\
    &\leq \frac{E\|(\hat{Q}^{-1} - Q) \mathbb{E}_n[p_i\epsilon_i] + Q\mathbb{E}_n[p_i\epsilon_i] \|}{t} \\
    &\leq \frac{E\|(\hat{Q}^{-1} - Q) \mathbb{E}_n[p_i\epsilon_i]\| + E\|\mathbb{E}_n[p_i\epsilon_i] \|}{t} \\
    &\lesssim \frac{(a+1) \sqrt{k/n}}{t}
\end{align}
\noindent where the $\sqrt{k/n}$ comes from
\begin{equation}
    E\|\mathbb{E}_n[p_i\epsilon_i] \|^2 = E[\sigma_i^2 p_i'p_i/n] \lesssim k/n,
\end{equation}
\noindent given that $\sigma_i^2 \leq \bar{\sigma}^2 < \infty $ and the $a$ term comes from applying the Rudelson and Vershynin result and the continuous mapping theorem for the inverse.

We proceed similarly for the $\|\hat{Q}^{-1} \mathbb{E}_n[p_ir_i]\|$ term by observing that
\begin{equation}
    \|\hat{Q}^{-1/2} \mathbb{E}_n[p_ir_i]\| \leq \mathbb{E}_n[r_i^2].
\end{equation}
Then, by Markov's inequality for $t>0$ when $c_k \to 0$
\begin{align}
    P(\|\hat{Q}^{-1} \mathbb{E}_n[p_ir_i]\| >t)
    &\leq \frac{E\|(\hat{Q}^{-1} - Q) \mathbb{E}_n[p_i r_i]\| + E\|\mathbb{E}_n[p_ir_i] \|}{t} \\
    &\lesssim \frac{(a+1) c_k}{t}
\end{align}
\noindent where the $c_k$ term comes from $E[r_i^2] \leq c_k^2$ and we note that all the eigenvalues of $\hat{Q}^{1/2}$ are bounded away from zero with high probability.

Combining both bounds yields the desired asymptotic result.

For the finite-sample rate, note that for some $\delta>0$,
\begin{align}
    P(\| \hat{\beta} - \beta\| > \delta) &\leq P(\|\hat{Q}^{-1} \mathbb{E}_n[p_i\epsilon_i]\| + \|\hat{Q}^{-1} \mathbb{E}_n[p_ir_i]\| > \delta) \\
    &\leq P(\|(\hat{Q}^{-1} - Q + Q)\mathbb{E}_n[p_i\epsilon_i]\| > \delta/2) + P(\|(\hat{Q}^{-1} - Q + Q)\mathbb{E}_n[p_ir_i]\| > \delta/2) \\
    &\leq P(\|\hat{Q}^{-1} - Q \|_{op}\|\mathbb{E}_n[p_i\epsilon_i]\| > \delta/4 ) + P(\|\hat{Q}^{-1} - Q\|_{op}\|\mathbb{E}_n[p_ir_i]\| > \delta/4)  \\
    &\quad +P(\|\mathbb{E}_n[p_i\epsilon_i]\|>\delta/4)+ P(\|\mathbb{E}_n[p_ir_i]\|>\delta/4) \\
    &\leq  2P(\|\hat{Q}^{-1} - Q \|_{op}> \sqrt{\delta}/2)  \\
    &\quad +P(\|\mathbb{E}_n[p_i\epsilon_i]\|>\sqrt{\delta}/2) +P(\|\mathbb{E}_n[p_ir_i]\|>\sqrt{\delta}/2) \\
    &\quad +P(\|\mathbb{E}_n[p_i\epsilon_i]\|>\delta/4)+ P(\|\mathbb{E}_n[p_ir_i]\|>\delta/4),
\end{align}
where the last inequality follows from $P(\{w | x(w) y(w) > t\}) \leq P(\{w | x(w) > \sqrt{t}\}\cap \{w | y(w) > \sqrt{t}\}) \leq P(\{w | x(w) > \sqrt{t}\}) +  P(\{w | y(w) > \sqrt{t}\})$. The first term can be bounded directly by applying \cite{rudelson2007sampling}:

\begin{equation}
    P(\|\hat{Q}^{-1} - Q \|_{op}> \sqrt{\delta}/2) \lesssim 2 \exp\left\{-c\frac{\delta}{4a^2}\right\},
\end{equation}
where $a$ is defined as above. For the terms that depend on $\epsilon_i$ let $S_n = \|\mathbb{E}_n[p_i\epsilon_i]\|$ and observe that under the assumption that $\epsilon_i$ is subGaussian and that $p_i$ is uniformly bounded we have that $S_n - ES_n$ is a centered subGaussian random variable with coefficient $\nu_i = \xi_k\sigma_i$. Then, for $\delta >0$ by a suitable Hoeffding's inequality
\begin{align}
    P(\|\mathbb{E}_n[p_i\epsilon_i]\|>\delta) &= P(S_n - ES_n + ES_n>\delta) \\
    &\leq P(S_n - ES_n >\delta/2)  + P(ES_n > \delta/2)\\
    &\leq \exp\left\{ -\frac{n \delta^2}{4\xi_k \bar{\sigma}}\right\},
\end{align}
where the second step requires that $ES_n \leq \sqrt{k/n} < \delta/2$ and $\bar{\sigma}$ could be included in the constant term. For the terms that depends on $r_i$ observe that $R_n = \| \mathbb{E}[p_ir_i]\|$ is a uniformly bounded random variable given our assumptions on $p$ and $r$. Therefore, we can show that $R_n - ER_n$ is a subGaussian random variable with variance factor $4nc_k\xi_k$. In Belloni et al. 2015 it is shown that $ER_n \lesssim \frac{c_k \xi_k}{\sqrt{n}}$. Hence, applying Hoeffding's inequality again
\begin{align}
    P(\|\mathbb{E}_n[p_ir_i]\|>\delta) &= P(R_n - ER_n + ER_n>\delta) \\
    &\leq P(R_n - ER_n >\delta/2)  + P(ER_n > \delta/2)\\
    &\lesssim \exp\left\{ -\frac{n\delta^2}{8\xi_k c_k}\right\},
\end{align}
where the second step requires that $\frac{c_k \xi_k}{\sqrt{n}} < \delta/2$. Grouping all constants and noting that for $\delta<1$, $\sqrt{\delta} > \delta$, we can derive the finite-sample bound by three exponential terms. It can be stated as follows, for $t\in(0,1)$, $c_k \leq t/2$, and $\frac{c_k \xi_k}{\sqrt{n}}, \sqrt{k/n} < t/8$,
\begin{align}
    P(\| \hat{g} - g\|_{F,2} > \delta) &\leq P(\|p'\hat{\beta} - p'\beta\|_{F,2} + c_k > \delta) \\
    &\leq P(\|p'\hat{\beta} - p'\beta\|_{F,2} > \delta/2) \\
    &\lesssim \exp\left\{-\frac{t}{a^2}\right\} + \exp\left\{ -\frac{nt^2}{\xi_k c_k}\right\} + \exp\left\{ -\frac{n t^2}{\xi_k}\right\}.
\end{align}
\end{proof}

\subsection{Proof of Corollary \ref{corollary:opt_rate_bpr}}
\begin{proof}
Under the appropriate normalization it is shown in \cite{newey1997convergence} that $\xi_{k}\lesssim k$ and in \cite{belloni2015} that $c_k \lesssim k^{-s/d}$ for the holder class of smoothness $s$ denoted by $\Sigma_s(\mathcal{X})$. Observe that the approximation bound is decreasing in $k$ as $s,d >0$, so when $k\asymp B(J)^{d/B}$ it follows that $c_k \lesssim B^{-s/d}( J)^{\lfloor-s/B\rfloor}$. Similarly, $\xi_k \lesssim B(J)^{\lfloor d/B\lfloor } \leq B(J)^{d/B}$.

Next, consider the conditions of Theorem 2 $c_k \leq t/2$ and $\frac{c_k \xi_k}{\sqrt{n}}\lor\sqrt{k/n} < t/8$. Given $c_k \leq t/2$, a sufficient inequality for the second condition to hold is $\xi_k < \sqrt{n}/4$. Hence, under the bound $\xi_{k}\lesssim k$ a sufficient condition to apply Theorem 2 given $c_k \leq t/2$ is $n > 16k^2$ (large enough $n$). To complete the proof then, it remains to show that the finite-sample bound in Theorem 2 achieves a unique minimum with respect to $B$ given the other parameters.

Consider the function $G$ as defined in the Corollary statement, given that $e^x$ is a strictly increasing convex function, it follows that $G$ is non-increasing in $B$ for any $n, d, J, s >0$ and $t>0$ such that $d\geq s$. Furthermore, the image of $G$ is contained in $[0,3]$ and $B$ belongs to a compact set in $\mathbb{Z}^+$ given by $\mathcal{B}(n,d,J,s,t,)=\{B\in \mathbb{R} | B^{-s/d}(J/B)^{-s} \leq t/2\}\cap \mathbb{Z}^+$. Therefore, for any $n, d, J$, $s$ and $t$ such that $d\geq s$, there exists a unique $B\in\mathcal{B}(n,d,J,s)$ that minimizes $G$ which coincides with $\sup \mathcal{B}(n,d,J,s)$. Rewriting the condition $d^{-s/d}(J)^{-s/B} \leq t/2$ we get that the desired result. This implies that $G$ is a well defined function with respect to $B$.
\end{proof}

\subsection{Proof of Theorem \ref{theorem:ME_rates}}
\begin{proof}
This proof  follows the proof of Theorem \ref{theorem:l_2_rates} but substituting $p(x)$ with $\nabla_j p(x)$.

Start by observing that 
\begin{equation}
    \| \partial_j \hat{g} - \partial_j g\|_{F,2} \leq \|(\nabla_j p)'\hat{\beta} - (\nabla_j p)'\beta\|_{F,2} + c'_k.
\end{equation}
Let $Q^{(j)} \equiv E[(\nabla_j p_i)(\nabla_j p_i)']$. Because $\nabla_j p_i$ is only non-zero for indices $\mathcal{S}_j$ the matrix $Q^{(j)}$ has zero entries in rows and columns in $\mathcal{S}_j^c$. Define $Q^{(j)}_{\mathcal{S_j}}$ as the matrix selection for rows and columns in $\mathcal{S}_j$ and conformably define $\beta_{\mathcal{S}_j}$ and $\hat{\beta}_{\mathcal{S}_j}$ as the parameter vectors for the $\mathcal{S}_j$ entries. It follows that 
\begin{align}
        \|(\nabla_j p)'(\hat{\beta} - \beta)\|_{F,2} &= \| Q^{(j)^{1/2}}(\hat{\beta} - \beta)\| \\ 
        &=\| Q^{(j)^{1/2}}_{\mathcal{S}_j}(\hat{\beta}_{\mathcal{S}_j} - \beta_{\mathcal{S}_j})\| \\
        &\leq \| Q^{(j)^{1/2}}_{\mathcal{S}_j}\|_{op} \|\hat{\beta}_{\mathcal{S}_j} - \beta_{\mathcal{S}_j}\| \\
        &\leq C_{\partial} \|\hat{\beta}_{\mathcal{S}_j} - \beta_{\mathcal{S}_j}\| .
\end{align}
To bound the estimation term, define the linear projection operator $P_j:\mathbb{R}^{k\times k} \to \mathbb{R}^{s_j\times s_j}$ that selects the $\mathcal{S}_j$ rows and columns. Applying the operator 
\begin{align}
    \|\hat{\beta}_{\mathcal{S}_j} - \beta_{\mathcal{S}_j}\| &= \|P_j(\hat{\beta}- \beta)\|\\
    &\leq \|P_j\hat{Q}^{-1} \mathbb{E}_n[p_i\epsilon_i]\| + \|P_j\hat{Q}^{-1} \mathbb{E}_n[p_ir_i]\|.
\end{align}
Because $P_j$ is a selection matrix its operator norm is 1. Therefore,
$$
\|P_j\hat{Q}^{-1}\| = \|P_j(\hat{Q}^{-1} - Q + Q)\| \leq  \|\hat{Q}^{-1} - Q\| + \|P_jQ\| = \|\hat{Q}^{-1} - Q\| + \|P_j\|,
$$
\noindent where we used the normalization $Q=I$ in the last step. Proceeding now as in the proof of Theorem 1, by Markov's inequality, for $t>0$

\begin{align}
    P(\|P_j\hat{Q}^{-1} \mathbb{E}_n[p_i\epsilon_i]\| >t) &\leq \frac{E\|P_j(\hat{Q}^{-1} - Q + Q) \mathbb{E}_n[p_i\epsilon_i]\|}{t} \\
    &\leq \frac{E\|P_j(\hat{Q}^{-1} - Q) \mathbb{E}_n[p_i\epsilon_i] + P_jQ\mathbb{E}_n[p_i\epsilon_i] \|}{t} \\
    &\leq \frac{E\|(\hat{Q}^{-1} - Q) \mathbb{E}_n[p_i\epsilon_i]\| + E\|P_j\mathbb{E}_n[p_i\epsilon_i] \|}{t} \\
    &\lesssim \frac{(a+1) \sqrt{s_j/n}}{t}
\end{align}
\noindent where the $\sqrt{s_j/n}$ comes from
\begin{equation}
    E\|P_j\mathbb{E}_n[p_i\epsilon_i] \|^2 = E[\sigma_i^2 p_{\mathcal{S}_j}'p_{\mathcal{S}_j}/n] \lesssim s_j/n,
\end{equation}
\noindent given that $\sigma_i^2 \leq \bar{\sigma}^2 < \infty $ and the $a$ term comes from applying the Rudelson and Vershynin result and the continuous mapping theorem for the inverse.

The bound for the term involving $r_i$ proceeds exactly as in Theorem \ref{theorem:l_2_rates} by noting that $\|P_j\hat{Q}^{-1} \mathbb{E}_n[p_ir_i]\|$ term by observing that
\begin{equation}
    \|\hat{Q}^{-1/2} \mathbb{E}_n[p_ir_i]\| \leq \mathbb{E}_n[r_i^2].
\end{equation}
Then, by Markov's inequality for $t>0$ when $c_k \to 0$

\begin{align}
    P(\|\hat{Q}^{-1} \mathbb{E}_n[p_ir_i]\| >t)
    &\leq \frac{E\|(\hat{Q}^{-1} - Q) \mathbb{E}_n[p_i r_i]\| + E\|\mathbb{E}_n[p_ir_i] \|}{t} \\
    &\lesssim \frac{(a+1) c_k}{t}
\end{align}
\noindent where the $c_k$ term comes from $E[r_i^2] \leq c_k^2$ and we note that all the eigenvalues of $\hat{Q}^{1/2}$ are bounded away from zero with high probability.

Combining both bounds yields the desired asymptotic result.
\end{proof}

\end{document}